# A Uniform Framework for Concept Definitions in Description Logics

**Giuseppe De Giacomo**                    DEGIACOMO@DIS.UNIROMA1.IT
*Università di Roma "La Sapienza"*
*Via Salaria 113, 00198 Roma, Italy*

**Maurizio Lenzerini**                    LENZERINI@DIS.UNIROMA1.IT
*Università di Roma "La Sapienza"*
*Via Salaria 113, 00198 Roma, Italy*

## Abstract

Most modern formalisms used in Databases and Artificial Intelligence for describing an application domain are based on the notions of class (or concept) and relationship among classes. One interesting feature of such formalisms is the possibility of defining a class, i.e., providing a set of properties that precisely characterize the instances of the class. Many recent articles point out that there are several ways of assigning a meaning to a class definition containing some sort of recursion. In this paper, we argue that, instead of choosing a single style of semantics, we achieve better results by adopting a formalism that allows for different semantics to coexist. We demonstrate the feasibility of our argument, by presenting a knowledge representation formalism, the description logic $\mu\mathcal{ALCQ}$, with the above characteristics. In addition to the constructs for conjunction, disjunction, negation, quantifiers, and qualified number restrictions, $\mu\mathcal{ALCQ}$ includes special fixpoint constructs to express (suitably interpreted) recursive definitions. These constructs enable the usual frame-based descriptions to be combined with definitions of recursive data structures such as directed acyclic graphs, lists, streams, etc. We establish several properties of $\mu\mathcal{ALCQ}$, including the decidability and the computational complexity of reasoning, by formulating a correspondence with a particular modal logic of programs called the modal mu-calculus.

## 1. Introduction

Most modern formalisms used in Databases and Artificial Intelligence for representing an application domain are based on the notions of class (or concept) and relationship among classes. For example, the object-oriented and semantics data models developed in Databases describe data in terms of classes (sometimes called entity types) and incorporate several features for establishing various forms of relationships between classes. On the other hand, the notion of class (often called concept or frame) and that of link among classes are provided in all structured formalisms for Knowledge Representation (frame-based languages, semantic networks, description logics, etc.). Finally, this notion is also present in several type systems of programming languages, specially those based on the object-oriented paradigm.

There are basically two ways of using and describing classes (concepts). In the first one, which we can call the *prescriptive approach*, the description formalism allows for expressing a number of properties of a class, thus prescribing constraints that the instances of the class must satisfy. In the second one, which we can call the *definitional approach*, the formalism allows for providing the *definition* of a class, i.e., a set of properties that *precisely character-* *ize* the instances of the class. While the prescriptive approach is quite well understood and established, the definitional approach is still the subject of an interesting debate, regarding both its nature and its semantic foundation. In particular, it is well known that there are various ways to assign a meaning to a class definition when it contains some sort of recursion (Baader, 1990, 1991; Nebel, 1991; Beneventano & Bergamaschi, 1992; Beeri, 1990).

In this paper, we are concerned with the semantic problems related to the definitional approach, arguing that, instead of choosing a single style of semantics for the knowledge representation formalism, we achieve better results by allowing different semantics to coexist.

We discuss this issue in the context of Description Logics[1], which are logics originally developed in Knowledge Representation to provide a formal reconstruction of frame-based languages. Description logics describe the domain of interest in terms of concepts, which represent classes of individuals, and roles, which are binary relations used to specify properties or attributes of individuals as well as links with other individuals (Nebel, 1990). Starting from atomic concepts, denoted simply by a name, more complex concepts are built by using a suitable set of constructs. For example, the expression *parent $\sqcap$ male $\sqcap$ $\forall$child.male* denotes the concept of father (male parent) whose children are all male. The symbol $\sqcap$ denotes the construct for *concept conjunction*, while $\forall$ denotes *universal role quantification*. Typically, concepts are structured into hierarchies determined by the properties associated with them. The hierarchical structure is defined in such a way that more specific concepts inherit the properties of the more general ones.

We introduce a description logic, called $\mu\mathcal{ALCQ}$, which extends the well-known description logic $\mathcal{ALC}$ (Schmidt-Schauß & Smolka, 1991) by including the so called *qualified number restrictions*, which are a very general form of cardinality constraints on roles, and special *fixpoint constructs*, which enable us to capture the various semantics for recursive definitions within a single formalism. Notably, the availability of these constructs makes it possible to combine the usual frame-based descriptions with definitions of recursive data structures such as directed acyclic graphs, lists, streams, etc.

We establish several properties of $\mu\mathcal{ALCQ}$, including the decidability and the computational complexity of reasoning, by formulating a correspondence with a particular modal logic of programs called the *modal mu-calculus*.

Recent articles, (e.g., Bergamaschi & Sartori, 1992; Borgida, 1992), advocate the use of description logics as a unifying framework for several types of database and knowledge representation formalisms. Indeed, it is possible to show that, depending on both the constructs and the semantics used, one can capture several database models and programming language type systems by using description logics. Therefore, the study presented in this paper is not merely confined to description logics, but is also applicable to other representation formalisms.

The paper is organized as follows. In Section 2, we present the basic notions regarding both description logics and fixpoints. In Section 3, we motivate our approach through a detailed discussion about the different semantics of concept definitions that have been considered in the literature, and we argue for a formalism in which the various semantics coexist. In Section 4, we present one such formalism, namely the logic $\mu\mathcal{ALCQ}$, and we discuss several of its properties. In Section 5 we study reasoning techniques for $\mu\mathcal{ALCQ}$ and expose the correspondence with modal mu-calculus. Finally, in Section 6, we draw the conclusions and discuss some open problems.

---

1. Also called Concept Languages or Terminological Languages.

## 2. Preliminaries

In this section, we briefly present the basic notions regarding both description logics, and fixpoints. The interested reader is referred to (Nebel, 1990) and (de Bakker, 1980) for a more complete introduction to the subjects.

### 2.1 Description Logics

Description logics allow one to represent a domain of interest in terms of *concepts* and *roles*. Concepts model classes of individuals, while roles model relationships between classes. Starting with atomic concepts (denoted by $A$) and atomic roles (denoted by $R$), which are concepts and roles described simply by a name, complex concepts and roles can be built by means of suitable constructs.

In this section, we concentrate on the description logic $\mathcal{ALCQ}$, obtained from the well-known description logic $\mathcal{ALC}$ (Schmidt-Schauß & Smolka, 1991) by including *qualified number restrictions*. These are cardinality constraints on the role fillers of a very general form, where role fillers to which a constraint applies are selected by means of a generic concept expression, the *qualifier*.

$\mathcal{ALCQ}$ concepts (denoted by $C$ or $D$, possibly with a subscript) are composed inductively according to the following *abstract syntax* ($n$ denotes a natural number):

$$C ::= A \mid \top \mid \bot \mid \neg C \mid C_1 \sqcap C_2 \mid C_1 \sqcup C_2 \mid \exists R.C \mid \forall R.C \mid (\leq n\,R.C) \mid (\geq n\,R.C).$$

These constructs are not all independent. The following equalities hold: $\top = A \sqcup \neg A$, $\bot = \neg\top$, $\forall R.C = \neg\exists R.\neg C$, and $(\leq n\,R.C) = \neg(\geq n+1\,R.C)$.

From a semantic point of view, concepts are interpreted as subsets of an abstract domain, while roles are interpreted as binary relations over such a domain. More precisely, an *interpretation* $\mathcal{I} = (\Delta^{\mathcal{I}}, \cdot^{\mathcal{I}})$ consists of a *domain of interpretation* $\Delta^{\mathcal{I}}$, and an *interpretation function* $\cdot^{\mathcal{I}}$ mapping every atomic concept $A$ to a subset of $\Delta^{\mathcal{I}}$ and every atomic role $R$ to a subset of $\Delta^{\mathcal{I}} \times \Delta^{\mathcal{I}}$.

The interpretation function $\cdot^{\mathcal{I}}$ is extended to complex concepts of $\mathcal{ALCQ}$ (note that in $\mathcal{ALCQ}$ roles are always atomic) as follows:

$$
\begin{aligned}
\top^{\mathcal{I}} &= \Delta^{\mathcal{I}} \\
\bot^{\mathcal{I}} &= \emptyset \\
(\neg C)^{\mathcal{I}} &= \Delta^{\mathcal{I}} - C^{\mathcal{I}} \\
(C_1 \sqcap C_2)^{\mathcal{I}} &= C_1^{\mathcal{I}} \cap C_2^{\mathcal{I}} \\
(C_1 \sqcup C_2)^{\mathcal{I}} &= C_1^{\mathcal{I}} \cup C_2^{\mathcal{I}} \\
(\exists R.C)^{\mathcal{I}} &= \{s \in \Delta^{\mathcal{I}} \mid \exists s'.\,(s,s') \in R^{\mathcal{I}} \text{ and } s' \in C^{\mathcal{I}}\} \\
(\forall R.C)^{\mathcal{I}} &= \{s \in \Delta^{\mathcal{I}} \mid \forall s'.\,(s,s') \in R^{\mathcal{I}} \text{ implies } s' \in C^{\mathcal{I}}\} \\
(\leq n\,R.C)^{\mathcal{I}} &= \{s \in \Delta^{\mathcal{I}} \mid \#\{s' \mid (s,s') \in R^{\mathcal{I}} \text{ and } s' \in C^{\mathcal{I}}\} \leq n\} \\
(\geq n\,R.C)^{\mathcal{I}} &= \{s \in \Delta^{\mathcal{I}} \mid \#\{s' \mid (s,s') \in R^{\mathcal{I}} \text{ and } s' \in C^{\mathcal{I}}\} \geq n\}
\end{aligned}
$$

where $\#S$ denotes the cardinality of the set $S$.

A *concept $C$ is satisfiable* iff there exists an interpretation $\mathcal{I}$ such that $C^{\mathcal{I}} \neq \emptyset$, otherwise $C$ is *unsatisfiable*. A *concept $C_1$ is subsumed by a concept $C_2$*, written as $C_1 \sqsubseteq C_1$, iff for every interpretation $\mathcal{I}$, $C_1^{\mathcal{I}} \subseteq C_2^{\mathcal{I}}$.

Our knowledge expressed in terms of concepts and roles is assembled into a special knowledge base, traditionally called *TBox*, which consists of a finite (possibly empty) set of assertions. In order to be as general as possible, we assume that every assertion has the form of an *inclusion assertion* (or simply inclusion):

$$C_1 \sqsubseteq C_2$$

without any restriction on the form of the concepts $C_1$ and $C_2$. A pair of inclusions of the form $\{C_1 \sqsubseteq C_2, C_2 \sqsubseteq C_1\}$ is often written as $C_1 \equiv C_2$ and is called *equivalence assertion*.

An interpretation $\mathcal{I}$ *satisfies* an inclusion $C_1 \sqsubseteq C_2$ iff $C_1^{\mathcal{I}} \subseteq C_2^{\mathcal{I}}$. An interpretation $\mathcal{I}$ is a *model of a TBox $\mathcal{K}$* iff $\mathcal{I}$ satisfies all inclusions in $\mathcal{K}$.

Let $\mathcal{K}$ be a TBox. We say that a *concept $C$ is satisfiable in $\mathcal{K}$*, iff there exists a model $\mathcal{I}$ of $\mathcal{K}$ such that $C^{\mathcal{I}} \neq \emptyset$, *unsatisfiable* otherwise. We say that a *concept $C_1$ is subsumed by a concept $C_2$ in $\mathcal{K}$*, written $\mathcal{K} \models C_1 \sqsubseteq C_2$, iff for every model $\mathcal{I}$ of $\mathcal{K}$, $C_1^{\mathcal{I}} \subseteq C_2^{\mathcal{I}}$.

## 2.2 Fixpoints

We briefly recall few notions on fixpoints. Consider the equation:

$$X = f(X)$$

where $f$ is an operator from $2^{\mathcal{S}}$ to $2^{\mathcal{S}}$ ($2^{\mathcal{S}}$ denotes the set of all subsets of a set $\mathcal{S}$). Every solution $\mathcal{E}$ of this equation is called a *fixpoint* of the operator $f$ (while every set $\mathcal{E}$ such that $f(\mathcal{E}) \subseteq \mathcal{E}$ is called *pre-fixpoint*, and every set $\mathcal{E}$ such that $\mathcal{E} \subseteq f(\mathcal{E})$ is called *post-fixpoint*). In general, an equation as the one above may have either no solution, a finite number of solutions, or an infinite number of them. Among the various solutions, the smallest and the greatest solutions (with respect to set-inclusion) have a prominent position, if they exist. A fundamental result due to Tarski (Tarski, 1955) guarantees the existence and the uniqueness of both such solutions in case $f$ is monotonic wrt set-inclusion ($\subseteq$), where $f$ is *monotonic* wrt $\subseteq$ whenever $\mathcal{E}_1 \subseteq \mathcal{E}_2$ implies $f(\mathcal{E}_1) \subseteq f(\mathcal{E}_2)$.

**Theorem 1 (Tarski)** *Let $\mathcal{S}$ be a set, and $f$ an operator from $2^{\mathcal{S}}$ to $2^{\mathcal{S}}$ that is monotonic wrt $\subseteq$. Then:*

- *There exists a unique least fixpoint of $f$, which is given by $\bigcap \{\mathcal{E} \subseteq \mathcal{S} \mid f(\mathcal{E}) \subseteq \mathcal{E}\}$.*

- *There exists a unique greatest fixpoint of $f$, which is given by $\bigcup \{\mathcal{E} \subseteq \mathcal{S} \mid \mathcal{E} \subseteq f(\mathcal{E})\}$.*

## 3. Concept Definitions as Equations

We now analyze the notion of concept definition in detail. Let us ignore for the moment knowledge bases as they have been introduced in the previous section, and let us consider a different kind of assertion: the definition statement. Let the form of a *definition statement* (or simply definition) be:

$$A =_{def} C$$

where $A$ is an atomic concept which cannot appear in the left-hand side of other definition statements, and $C$ is a concept expression in $\mathcal{ALCQ}$. In principle, $A =_{def} C$ is intended to provide an exact account for the concept to $A$ in terms of $C$, i.e., to define $A$ as *the* set of the individuals satisfying $C$.

In specifying the semantics of definitions, we need to distinguish between two different types of atomic concepts, namely, *primitive concepts* and *defined concepts*. Given a set of definition statements, the primitive concepts are the atomic concepts that do not appear in the left-hand side of any definition statement, whereas the defined concepts are those that appear in the left-hand side of a definition statement.

Given an interpretation $\mathcal{I} = (\Delta^{\mathcal{I}}, \cdot^{\mathcal{I}})$, the interpretation function $\cdot^{\mathcal{I}}$ directly assigns a subset of $\Delta^{\mathcal{I}}$ to primitive concepts, but not to defined concepts. The meaning of a defined concept $A$ is assigned through its definition statement $A =_{def} C$, extending the interpretation function so as the following requirement is satisfied:

$$A^{\mathcal{I}} = C^{\mathcal{I}}. \tag{1}$$

Consider, for example, the definition statement:

$$parent =_{def} \exists child.\top.$$

Note that the defined concept *parent* does not appear in the body of its definition statement. By (1), the definition statement provides the *definition* for the concept *parent*, in the following sense: in any interpretation $\mathcal{I} = (\Delta^{\mathcal{I}}, \cdot^{\mathcal{I}})$, $parent^{\mathcal{I}}$ denotes a *single* subset of $\Delta^{\mathcal{I}}$, exactly the one denoted by $(\exists child.\top)^{\mathcal{I}}$, i.e., $\{s \mid \exists t.(s,t) \in child^{\mathcal{I}}\}$. In general, if a concept $A$ is defined in terms of primitive and already defined concepts, then for every interpretation $\mathcal{I}$ there exists a unique way to extend the interpretation function to defined concepts, and hence there is no doubt that the definition statement provides a *definition* of $A$.

Now, consider the following definition statement:

$$A =_{def} \exists child.A.$$

Given an interpretation $\mathcal{I} = (\Delta^{\mathcal{I}}, \cdot^{\mathcal{I}})$, by (1) the statement is interpreted as the equation:

$$A^{\mathcal{I}} = \{s \in \Delta^{\mathcal{I}} \mid \exists t.(s,t) \in child^{\mathcal{I}} \text{ and } t \in A^{\mathcal{I}}\}.$$

However such equation does not specify univocally how to extend the interpretation function $\cdot^{\mathcal{I}}$ to the defined concept $A$, since $\emptyset$ satisfies the equation as well as any set of individuals where each member has an infinite chain of descendants that are also members.

In general, we call *recursive definition statements*[2] (or simply recursive definitions), definition statements of the form:

$$A =_{def} F(A)$$

where $F(A)$ stands for a concept that has $A$ as a subconcept[3]. According to (1), the recursive definition $A =_{def} F(A)$ is interpreted simply as a sort of equation specifying that, given an interpretation $\mathcal{I}$, the subset of $\Delta^{\mathcal{I}}$ that can be tied to the defined concept $A$ must satisfy the equation $A^{\mathcal{I}} = F(A)^{\mathcal{I}}$, i.e., must be one of its *solutions*. Observe that, in general, either none, one, or several subsets of $\Delta^{\mathcal{I}}$ may exist which are solutions of the above equation.

---

2. Terminological cycles in (Baader, 1990, 1991; Nebel, 1991). In the present discussion, for sake of simplicity, we do not consider mutual recursive definitions, as $A =_{def} F(B)$, $B =_{def} F'(A)$. We shall come back to this point later on.

3. A *subconcept* of a concept $C$ is any substring of $C$ (including $C$ itself) that is a concept, according to the syntax rules.

Another convenient way to consider a definition statement is to associate to it, for every interpretation $\mathcal{I}$, an operator from subsets of $\Delta^{\mathcal{I}}$ to subsets of $\Delta^{\mathcal{I}}$ instead of an equation, so that the fixpoints of the operator correspond to the solutions of the equation. For example, to the definition $A =_{def} \exists child.A$ we associate the operator:

$$\lambda S.\{s \in \Delta^{\mathcal{I}} \mid \exists t.(s,t) \in child^{\mathcal{I}} \text{ and } t \in S\}$$

for any interpretation $\mathcal{I}$. In general as either none, one or multiple solutions exist for the equation associated with a recursive definition, we have that either none, one or multiple fixpoints exist for the corresponding operator.

In this situation the word "definition" itself seems misleading: the body of the definition does not give a complete account of the defined concept. An additional criterion is needed for selecting solutions of the associated equation, or equivalently, fixpoints of the associated operator. In other words in addition to (1), a criterion is needed to extend univocally the interpretations $\mathcal{I}$ to the defined concepts. This observation has led to various semantics, each of which interprets recursive definitions differently, by choosing, a priori and once and for all, which solutions, or equivalently which fixpoints, are to be assigned to the defining concept of a recursive definition[4].

## 3.1 Different Semantics for Recursive Definitions

In the literature on description logics, three semantics for recursive definitions have been proposed (see Nebel, 1991):

- Descriptive Semantics

- Least Fixpoint Semantics

- Greatest Fixpoint Semantics

and which of these semantics is the "right" one is a long standing matter of debate. Below we describe how each of the three semantics interprets recursive definitions, and present some examples showing that the choice of the semantics depends in fact upon the concept to be defined. But first, it should be stressed that only the descriptive semantics is able to assign meaning to general inclusion assertions $C_1 \sqsubseteq C_2$ introduced in the previous section.

According to the *Descriptive Semantics*, a recursive definition $A =_{def} F(A)$ is a *constraint* stating that, for any $\mathcal{I}$, $A^{\mathcal{I}}$ has to be *a* solution of the equation $A^{\mathcal{I}} = F(A)^{\mathcal{I}}$. Under the descriptive semantics, $A =_{def} \exists child.A$ simply states that the individuals in the class $A$ have a child in the class $A$, and the individuals that have a child in the class $A$ are themselves in the class $A$, where $A$ *is no better specified*. In general the descriptive semantics is not appropriate to properly define recursive concepts, in the sense that, given an interpretation $\mathcal{I}$, it is unable to assign a unique subset of $\Delta^{\mathcal{I}}$ to the defined concept of the recursive definition.

---

4. We remark that a non-recursive definition is interpreted by the various semantics in the same way, since, for every $\mathcal{I}$, the equation associated to it has a single solution.

In fact under descriptive semantics definition statements are indistinguishable from the equivalence assertions introduced in the previous section. In other words, the meaning assigned to $A =_{def} F(A)$ is the same as that assigned to the equivalence assertion $A \equiv F(A)$. Although such equivalence assertions can be used to specify if-and-only-if constraints, they do not provide proper definitions when recursion is involved. For example, we can express the fact that humans are mammals having parents that are humans, and on the converse, that mammals having parents that are humans are humans themselves, in terms of the equivalence assertion:

$$human \equiv mammal \sqcap \exists parent.\top \sqcap \forall parent.human.$$

Similarly we may require horses to satisfy an analogous property:

$$horse \equiv mammal \sqcap \exists parent.\top \sqcap \forall parent.horse.$$

However the two equivalence assertions above do not define *human* and *horse* as shown, e.g., by the fact that (correctly) they do not imply that all humans are horses and vice-versa (in contrast to what happen when a fixpoint semantics is used, see below).

The *Least Fixpoint Semantics* interprets a recursive definition statement $A =_{def} F(A)$ by assigning to $A$ the smallest possible extension in each interpretation $\mathcal{I}$, among those that satisfy $A^{\mathcal{I}} =_{def} F(A)^{\mathcal{I}}$ – i.e., the least fixpoint of the corresponding operator. In fact it is always assumed that the operator associated with the definition statement is monotonic, so that Theorem 1 applies and a least fixpoint exists and is unique, i.e., the corresponding equation has a *unique* smallest solution. Hence under the least fixpoint semantics the recursive definition statement $A =_{def} F(A)$ *defines* the concept $A$. It is easy to verify that in the example $A =_{def} \exists child.A$, the least fixpoint semantics leads us to identify $A$ with $\bot$. Indeed the empty set is a solution of the equation associated with the statement, and it is obviously the smallest solution. Similarly if we interpret the definition statement:

$$human =_{def} mammal \sqcap \exists parent.\top \sqcap \forall parent.human$$

with the least fixpoint semantics, we have that $human^{\mathcal{I}} = \emptyset$ for any interpretation $\mathcal{I}$. Observe that if, as above, we adopt a similar definition for *horse*, we get again $horse^{\mathcal{I}} = \emptyset$, so we can trivially infer that $horse \equiv human$.

The least fixpoint semantics is particularly suitable for providing *inductive definitions* of concepts. For example, consider the class of a list (LIST) defined as follows:

- An EMPTY-LIST is a LIST.

- A NODE that has exactly one successor that is a LIST is a LIST.

- Nothing else is a LIST.

The first two conditions can be captured by the following recursive definition statement[5]:

$$list =_{def} emptylist \sqcup (node \sqcap (\leq 1\, succ.\top) \sqcap \exists succ.list)$$

---

5. Additionally we specify that the two concepts *emptylist* and *node* are disjoint.

where $(\leq 1\, succ.\top)$ forces $succ$ to be a function. To enforce the third condition we must assign the smallest possible extension to $list$. Thus, the class of LISTs can be naturally defined by means of the above definition statement, interpreted according to the least fixpoint semantics.

The *Greatest Fixpoint Semantics* interprets a recursive definition statement $A =_{def} F(A)$ by assigning to $A$ the largest possible extension in each interpretation $\mathcal{I}$, among those that satisfy $A^{\mathcal{I}} =_{def} F(A)^{\mathcal{I}}$ – i.e., the greatest fixpoint of the corresponding operator. Again, it is assumed that such operator is monotonic in order to guaranty the existence an the unicity of the greatest fixpoint (Theorem 1). As for the least fixpoint semantics, under the greatest fixpoint semantics a recursive definition statement $A =_{def} F(A)$ *defines* the concept $A$. For example, considering again the definition statement $A =_{def} \exists child.A$, the greatest fixpoint semantics leads us to interpret $A$ as the class of *all* the individuals having a child that in turn is a member of $A$.

While the least fixpoint semantics naturally captures classes defined by induction, the greatest fixpoint semantics naturally captures classes of individuals whose structure is *non-well-founded* or *co-inductive*. An example is the class of STREAMs, modeling the well-known linear data structure having a NODE as first element, and such that the rest of the structure is a STREAM itself. Note that streams are similar to lists except that while lists can be considered as finite sequences of nodes, streams are infinite sequences of nodes. Such a class can be captured by the following recursive definition statement:

$$stream =_{def} node \sqcap (\leq 1\, succ.\top) \sqcap \exists succ.stream$$

with the proviso that the greatest possible extension is assigned to $stream$.

Finally, consider under the greatest fixpoint semantics the recursive definition statements:

$$human =_{def} mammal \sqcap \exists parent.\top \sqcap \forall parent.human$$

$$horse =_{def} mammal \sqcap \exists parent.\top \sqcap \forall parent.horse.$$

Although they do not assign the empty extension to both *human* and *horse* as the least fixpoint semantics does, we do have again the rather counter intuitive consequence that $human \equiv horse$, since $human^{\mathcal{I}} = horse^{\mathcal{I}}$ for any interpretation $\mathcal{I}$. In general under both types of fixpoint semantics the particular name used to denote a defined concept does not have any impact on the interpretation of it, since the meaning of the defined concept is completely specified by its definition statement.

## 3.2 Least and Greatest Fixpoints as Concept Constructs

The above considerations show that arguing about which is the "right" semantics for recursive definitions is not really an issue. Each of them captures an essential use of recursive equations: the descriptive semantics is appropriate to specify constraints on concepts and is the only one that extends to the general inclusion assertions introduced in Section 2; the least fixpoint semantics is appropriate to define a structure inductively; the greatest fixpoint semantics is the appropriate to define non-well-founded structures. Generally, we may need the three of them in the same knowledge base, in order to model the various properties of the different concepts.

Our proposal in this paper is exactly in the direction of reconciling the various semantics in the same knowledge base. This is pursued by means of a logic that incorporates two constructs, $\mu X.F(X)$ and $\nu X.F(X)$ (the symbols $X, Y, \ldots$ stand for concept variables), denoting, respectively, the least fixpoint and the greatest fixpoint of the operator associated with the definition $X =_{def} F(X)$, that is, for every $\mathcal{I}$, the smallest solution and the greatest solution of the equation $X^{\mathcal{I}} = F(X)^{\mathcal{I}}$.

In our approach, definition statements will never appear in a TBox. Instead, a knowledge base will be simply a set of inclusion assertions (interpreted according to the descriptive semantics) that can involve fixpoint constructs. For example, in order to specify the properties of the concepts of *list*, *stream*, *human* and *horse* we can use the following assertions[6]:

$$
\begin{array}{rcl}
list & \equiv & \mu X. emptylist \sqcup (node \sqcap (\leq 1\, succ.\top) \sqcap \exists succ.X) \\
stream & \equiv & \nu X. node \sqcap (\leq 1\, succ.\top) \sqcap \exists succ.X \\
human & \equiv & mammal \sqcap \exists parent.\top \sqcap \forall parent.human \\
horse & \equiv & mammal \sqcap \exists parent.\top \sqcap \forall parent.horse.
\end{array}
$$

Note that, if we then add to the above knowledge base the equivalence assertion:

$$mgm \equiv \nu X\,.\, mammal \sqcap \exists parent.\top \sqcap \forall parent.X$$

defining the concept *mgm* (mammal generated by mammal), then it correctly turns out that both *human* and *horse* are subsumed by *mgm*.

The availability of least and greatest fixpoint constructs, by allowing different semantics to be used in the same TBox, makes it possible to model not only abstract classes, but also inductively and co-inductively defined data structures, such as lists and streams. This is particularly important if our objective is to integrate class-based representation formalisms and programming systems (declarative or procedural), in order to make these formalisms more useful in practice. Furthermore, we have the possibility of nesting fixpoints, thus going beyond the simple equational format by which we motivated their introduction. As an example, consider the following one:

> Among the inhabitants of the planet "Plonk", a disease called "foo" is quite common. Such a disease manifests itself in two forms: a "visible" one and a "latent" one, and it has a rather intricate hereditary pattern. Individuals that have the visible form transmit the visible form to at least one (say the first) direct descendant (obviously, if there is any direct descendant), these ill descendants in turn do the same, and so on, *until* someone transmits the latent form of the disease. More precisely, along any chain of descendants, the visible form of the disease sooner or later is interrupted, because either an individual has no direct descendant or an individual transmits to some descendant the latent form. All direct descendants (if any) of an individual that has the latent form inherit the visible form. The pattern goes on like this, generation after generation, *forever*.

The hereditary pattern (*foo_hp*) of the above disease can be defined as follows:

$$
\begin{array}{l}
foo\_hp \equiv \nu X.\mu Y.((visible \sqcap (\exists child.Y \sqcup \forall child.\bot)) \sqcup \\
\qquad\qquad (latent \sqcap \forall child.(visible \sqcap X)))
\end{array}
$$

---

6. We also include the assertion $emptylist \sqsubseteq \neg node$, specifying that the concepts *emptylist* and *node* are disjoint.

where *visible* and *latent* denote the visible and the latent form respectively of the disease, and are assumed to be disjoint (*latent* $\sqsubseteq \neg visible$).

## 4. The Description Logic $\mu\mathcal{ALCQ}$

We provide a formal account of the meaning of the fixpoint constructs by introducing a description logic, called $\mu\mathcal{ALCQ}$, which is obtained by adding these constructs to $\mathcal{ALCQ}$.

We make use of the notions of scope, bound and free occurrences of variables, closed formulae, etc. The definitions of these notions are the same as the analogues in first-order logic, treating $\mu$ and $\nu$ as quantifiers. In addition, we use the symbol $\sigma$ as an abstraction for either $\mu$ or $\nu$.

The primitive symbols in $\mu\mathcal{ALCQ}$ are atomic concepts, *(concept) variables* (denoted by $X, Y, \ldots$), and atomic roles which are the only roles admitted in the logic.

Concepts in $\mu\mathcal{ALCQ}$ are formed inductively according to the following abstract syntax:

$$C \quad ::= \quad A \mid \top \mid \bot \mid \neg C \mid C_1 \sqcap C_2 \mid C_1 \sqcup C_2 \mid \exists R.C \mid \forall R.C \mid (\leq n\,R.C) \mid (\geq n\,R.C) \mid$$
$$\mu X.C \mid \nu X.C \mid X$$

where $A$ denotes an atomic concept, $R$ an atomic role, $n$ a natural number, and $X$ a variable, and *with the restriction that only a variable $X$ occurring positively in $C$ can be bounded by a fixpoint $\sigma$ in $\sigma X.C$*. We say that a variable $X$ occurs *positively* in a concept $C$, if every free occurrence of $X$ is in the scope of an even number of negations, considering concepts $C'$ in $(\leq n\,R.C')$ in the scope of one negation.

The two fixpoint constructs are mutually definable: $\nu X.C = \neg\mu X.\neg C[X/\neg X]$ (where $C[X/\neg X]$ is the concept obtained substituting all free occurrences of $X$ by the concept $\neg X$).

As before, an interpretation $\mathcal{I} = (\Delta^{\mathcal{I}}, \cdot^{\mathcal{I}})$ consists of a domain of interpretation $\Delta^{\mathcal{I}}$, and a interpretation function $\cdot^{\mathcal{I}}$, which maps every atomic concept to a subset of $\Delta^{\mathcal{I}}$, and every atomic role to a subset of $\Delta^{\mathcal{I}} \times \Delta^{\mathcal{I}}$. But the presence of free variables does not allow us to extend the interpretation function $\cdot^{\mathcal{I}}$ directly to every concept of the logic. For this reason we introduce valuations. A *valuation* $\rho$ on an interpretation $\mathcal{I}$ is a mapping from variables to subsets of $\Delta^{\mathcal{I}}$.

Given a valuation $\rho$, we denote by $\rho[X/\mathcal{E}]$ the valuation identical to $\rho$ except for $\rho[X/\mathcal{E}](X) = \mathcal{E}$. In other words, for every variable $Y$:

$$\rho[X/\mathcal{E}](Y) = \begin{cases} \mathcal{E} & \text{if } Y = X \\ \rho(Y) & \text{if } Y \neq X \end{cases}$$

Let $\mathcal{I}$ be an interpretation and $\rho$ a valuation on $\mathcal{I}$. We assign meaning to concepts of the logic by associating to $\mathcal{I}$ and $\rho$ an *extension function* $\cdot^{\mathcal{I}}_{\rho}$, mapping concepts to subsets of $\Delta^{\mathcal{I}}$, as follows:

$$
\begin{aligned}
X_\rho^{\mathcal{I}} &= \rho(X) \subseteq \Delta^{\mathcal{I}} \\
A_\rho^{\mathcal{I}} &= A^{\mathcal{I}} \subseteq \Delta^{\mathcal{I}} \\
\top_\rho^{\mathcal{I}} &= \Delta^{\mathcal{I}} \\
\bot_\rho^{\mathcal{I}} &= \emptyset \\
(\neg C)_\rho^{\mathcal{I}} &= \Delta^{\mathcal{I}} - C_\rho^{\mathcal{I}} \\
(C_1 \sqcap C_2)_\rho^{\mathcal{I}} &= (C_1)_\rho^{\mathcal{I}} \cap (C_2)_\rho^{\mathcal{I}} \\
(C_1 \sqcup C_2)_\rho^{\mathcal{I}} &= (C_1)_\rho^{\mathcal{I}} \cup (C_2)_\rho^{\mathcal{I}} \\
(\exists R.C)_\rho^{\mathcal{I}} &= \{ s \in \Delta^{\mathcal{I}} \mid \exists s'.\ (s, s') \in R^{\mathcal{I}} \text{ and } s' \in C_\rho^{\mathcal{I}} \} \\
(\forall R.C)_\rho^{\mathcal{I}} &= \{ s \in \Delta^{\mathcal{I}} \mid \forall s'.\ (s, s') \in R^{\mathcal{I}} \text{ implies } s' \in C_\rho^{\mathcal{I}} \} \\
(\leq n\, R.C)_\rho^{\mathcal{I}} &= \{ s \in \Delta^{\mathcal{I}} \mid \#\{ s' \mid (s, s') \in R^{\mathcal{I}} \text{ and } s' \in C_\rho^{\mathcal{I}} \} \leq n \} \\
(\geq n\, R.C)_\rho^{\mathcal{I}} &= \{ s \in \Delta^{\mathcal{I}} \mid \#\{ s' \mid (s, s') \in R^{\mathcal{I}} \text{ and } s' \in C_\rho^{\mathcal{I}} \} \geq n \} \\
(\mu X.C)_\rho^{\mathcal{I}} &= \bigcap \{ \mathcal{E} \subseteq \Delta^{\mathcal{I}} \mid \quad C_{\rho[X/\mathcal{E}]}^{\mathcal{I}} \subseteq \mathcal{E} \} \\
(\nu X.C)_\rho^{\mathcal{I}} &= \bigcup \{ \mathcal{E} \subseteq \Delta^{\mathcal{I}} \mid \quad \mathcal{E} \subseteq C_{\rho[X/\mathcal{E}]}^{\mathcal{I}} \}
\end{aligned}
$$

In the last two cases $C_{\rho[X/\mathcal{E}]}^{\mathcal{I}}$ is interpreted as an operator from subsets $\mathcal{E}$ of $\Delta^{\mathcal{I}}$ to subsets of $\Delta^{\mathcal{I}}$. By the syntactic restriction enforced on variables, such an operator is guaranteed to be monotonic wrt $\subseteq$. Notice that free variables appearing in a concept are interpreted similarly to atomic concepts.

A concept $C$ is satisfiable, if there exists an interpretation $\mathcal{I}$ and a valuation $\rho$ on $\mathcal{I}$ such that $C_\rho^{\mathcal{I}} \neq \emptyset$, otherwise $C$ is unsatisfiable. A concept $C_1$ is subsumed by a concept $C_2$, written as $C_1 \sqsubseteq C_2$, if for every interpretation $\mathcal{I}$ and every valuation $\rho$ on $\mathcal{I}$, $(C_1)_\rho^{\mathcal{I}} \subseteq (C_2)_\rho^{\mathcal{I}}$.

A $\mu\mathcal{ALCQ}$ TBox is a finite (possibly empty) set of inclusion assertions $C_1 \sqsubseteq C_2$ where $C_1$ and $C_2$ are *closed* concepts of $\mu\mathcal{ALCQ}$. As before, we use equivalence assertions of the form $C_1 \equiv C_2$ as an abbreviation for $\{ C_1 \sqsubseteq C_2, C_2 \sqsubseteq C_1 \}$.

An interpretation $\mathcal{I}$ satisfies an inclusion assertion $C_1 \sqsubseteq C_2$, if $(C_1)_\rho^{\mathcal{I}} \subseteq (C_2)_\rho^{\mathcal{I}}$, where $\rho$ is any valuation on $\mathcal{I}$ (being $C_1$ and $C_2$ closed, and hence independent from valuations). $\mathcal{I}$ is a model of a TBox $\mathcal{K}$, if $\mathcal{I}$ satisfies all inclusion assertions in $\mathcal{K}$. We say that a TBox $\mathcal{K}$ is satisfiable, if it has a model. Observe that inclusion assertions in $\mathcal{K}$ are interpreted according to the descriptive semantics.

We say that a TBox $\mathcal{K}$ logically implies an inclusion assertion $C_1 \sqsubseteq C_2$, written $\mathcal{K} \models C_1 \sqsubseteq C_2$, if for every model $\mathcal{I}$ of $\mathcal{K}$ and every valuation $\rho$ on $\mathcal{I}$, $(C_1)_\rho^{\mathcal{I}} \subseteq (C_2)_\rho^{\mathcal{I}}$.

## 4.1 Properties of the Fixpoint Constructs

In the following, we use the notation $C(X)$ to indicate that the variable $X$ occurs free in the concept $C$ (other variables could occur free in $C$ as well), and the notation $C(D)$, where $D$ is a concept, as a shorthand for $C(X)[X/D]$ (i.e., the concept obtained substituting all free occurrences of $X$ in $C(X)$ by the concept $D$).

Let us comment briefly on some simple properties of the logic. First, the concept $\sigma X.C(X)$ is equivalent to the concept $\sigma Y.C(Y)$, as long as $Y$ is free for $X$ in $C(X)$. Second, the extension function $\cdot_\rho^{\mathcal{I}}$ assign to a closed concept a value which is independent of the actual valuation $\rho$. Hence $\sigma X.C$, where $X$ does not occur in $C$, is equivalent to $C$. Third, since $\sigma X.C(X)$ is a fixpoint we have that $C(\sigma X.C(X))$ is equivalent to $\sigma X.C(X)$. Furthermore, we have that the concept $\mu X.C(X)$ is always subsumed by the concept $\nu X.C(X)$.

The next property is more substantial. Consider the class of a single source finite directed acyclic graphs (DAGs) defined inductively as follows[7]:

- The EMPTY-DAG is a DAG (base step).

- A NODE that has connections and all connections are DAGs, is a DAG (inductive step).

- Nothing else is a DAG.

Consider now a $\mu\mathcal{ALCQ}$ TBox $\mathcal{K}$ containing the two equivalence assertions:

$$dag\_of\_student \equiv \mu X \,.\, emptydag \sqcup (student \sqcap \exists arc.\top \sqcap \forall arc.X)$$

$$dag\_of\_person \equiv \mu X \,.\, emptydag \sqcup (person \sqcap \exists arc.\top \sqcap \forall arc.X)$$

which define the concepts $dag\_of\_student$ and $dag\_of\_person$ as the classes of DAGs whose nodes are students and persons respectively. Assuming that students are persons, we want to be able to infer that DAGs of students are DAGs of persons as well. That is, we want:

$$\mathcal{K} \models student \sqsubseteq person \;\; \text{implies} \;\; \mathcal{K} \models dag\_of\_student \sqsubseteq dag\_of\_person.$$

It turns out that for $\mu\mathcal{ALCQ}$ such a property holds. To prove this, we introduce the following two theorems.

**Theorem 2** *Let $\mathcal{K}$ be a $\mu\mathcal{ALCQ}$ TBox, and $C$ and $D$ two $\mu\mathcal{ALCQ}$ concepts in which a variable $X$ occurs free positively. Then:*

$$\mathcal{K} \models C \sqsubseteq D \quad implies \quad \mathcal{K} \models \sigma X.C \sqsubseteq \sigma X.D.$$

**Proof** We proceed by contradiction[8]. Assume that $C_\rho^{\mathcal{I}} \subseteq D_\rho^{\mathcal{I}}$ holds for all models $\mathcal{I}$ of $\mathcal{K}$ and all valuations $\rho$ on $\mathcal{I}$. And suppose that there exists a model $\mathcal{I}$ of $\mathcal{K}$ and a valuation $\rho$ on $\mathcal{I}$ such that $(\sigma X.C)_\rho^{\mathcal{I}} \not\subseteq (\sigma X.D)_\rho^{\mathcal{I}}$.

First we prove the result for $\sigma = \mu$. Let $s$ be an individual in $(\mu X.C)_\rho^{\mathcal{I}}$ but not in $(\mu X.D)_\rho^{\mathcal{I}}$. Now, we have:

$$s \in (\mu X.C)_\rho^{\mathcal{I}} \;\; \text{iff} \;\; \forall \mathcal{E} \subseteq \Delta^{\mathcal{I}} \,.\, (C_{\rho[X/\mathcal{E}]}^{\mathcal{I}} \subseteq \mathcal{E} \text{ implies } s \in \mathcal{E}) \tag{2}$$

$$s \notin (\mu X.D)_\rho^{\mathcal{I}} \;\; \text{iff} \;\; \exists \mathcal{E}' \subseteq \Delta^{\mathcal{I}} \,.\, (D_{\rho[X/\mathcal{E}']}^{\mathcal{I}} \subseteq \mathcal{E}' \text{ and } s \notin \mathcal{E}'). \tag{3}$$

For the set $\mathcal{E}'$ in (3), the following expression holds:

$$C_{\rho[X/\mathcal{E}']}^{\mathcal{I}} \subseteq D_{\rho[X/\mathcal{E}']}^{\mathcal{I}} \subseteq \mathcal{E}'$$

---

7. We assume that a leaf of a DAG is a NODE with all arcs leading to a special DAG called EMPTY-DAG. As an alternative, one can assume that a leaf of a DAG is a NODE having no connection at all. In the latter case, the definition of $dag$ would simplify to $dag =_{def} node \sqcap \forall arc.dag$ (in which the general form of inductive definitions – i.e., base case and inductive case – is less apparent).

8. For uniformity, we do not distinguish if $X$ occurs free or not. Obviously if $X$ does not occur free, the result is trivial.

hence by (2) we have $s \in \mathcal{E}'$ and by (3) we have $s \notin \mathcal{E}'$, which is impossible.

The proof for $\sigma = \nu$ is similar. Let $s$ be an individual in $(\nu X.C)^{\mathcal{I}}_{\rho}$ but not in $(\nu X.D)^{\mathcal{I}}_{\rho}$. Now, we have:

$$s \in (\nu X.C)^{\mathcal{I}}_{\rho} \text{ iff } \exists \mathcal{E}'' \subseteq \Delta^{\mathcal{I}}. \ (\mathcal{E}'' \subseteq C^{\mathcal{I}}_{\rho[X/\mathcal{E}'']} \text{ and } s \in \mathcal{E}'') \qquad (4)$$

$$s \notin (\nu X.D)^{\mathcal{I}}_{\rho} \text{ iff } \forall \mathcal{E} \subseteq \Delta^{\mathcal{I}}. \ (\mathcal{E} \subseteq D^{\mathcal{I}}_{\rho[X/\mathcal{E}]} \text{ implies } s \notin \mathcal{E}). \qquad (5)$$

For the set $\mathcal{E}''$ in (4), the following expression holds:

$$\mathcal{E}'' \subseteq C^{\mathcal{I}}_{\rho[X/\mathcal{E}'']} \subseteq D^{\mathcal{I}}_{\rho[X/\mathcal{E}'']}$$

hence by (4) we have $s \in \mathcal{E}''$ and by (5) we have $s \notin \mathcal{E}''$, which is impossible. □

Above we have defined what it means for a variable $X$ to occur positively in a concept $C$. Similarly we say that a variable $X$ occurs *negatively* in a concept $C$, if every free occurrence of $X$ is in the scope of an odd number of negations, considering concepts $C'$ in $(\leq n R.C')$ in the scope of one negation.

**Theorem 3** *Let $\mathcal{K}$ be a $\mu\mathcal{ALCQ}$ TBox, and $D(X)$ a $\mu\mathcal{ALCQ}$ concept with the variable $X$ as a free variable. Then, for any $\mu\mathcal{ALCQ}$ concepts $C_1$ and $C_2$:*

$$\mathcal{K} \models C_1 \sqsubseteq C_2 \text{ implies } \begin{cases} \mathcal{K} \models D(C_1) \sqsubseteq D(C_2) & \text{if } X \text{ occurs positively in } D(X) \\ \mathcal{K} \models D(C_2) \sqsubseteq D(C_1) & \text{if } X \text{ occurs negatively in } D(X) \end{cases}$$

**Proof** We prove the result by induction on the formation of $D(X)$.

Base case. If $D(X) = X$, the result holds trivially.

Inductive cases. If $D(X)$ has the form $\neg D'(X) \mid (\geq n R.C')$ , then $X$ occurs positively (negatively) in $D'(X)$ and negatively (positively) in $D(X)$. By induction hypothesis $\mathcal{K} \models D'(C_i) \sqsubseteq D'(C_j)$ (where $i, j \in \{1, 2\}$ and $i \neq j$) and hence by the semantics of the constructs $\mathcal{K} \models D(C_j) \sqsubseteq D(C_i)$.

If $D(X)$ has the form $D'_1(X) \sqcap D'_2(X) \mid D'_1(X) \sqcup D'_2(X) \mid \forall R.D'(X) \mid (\geq n R.D'(X))$, then $X$ occurs positively (negatively) both in $D'(X)$ and in $D(X)$. By induction hypothesis $\mathcal{K} \models D'(C_i) \sqsubseteq D'(C_j)$ and hence by the semantics of the constructs $\mathcal{K} \models D(C_i) \sqsubseteq D(C_j)$.

It remains to prove the result for $D(X) = \sigma Y.D'(X) \ (Y \neq X)$. In this case, by the syntactic restriction enforced, $Y$ occurs positively in $D'(X)$ and hence by Theorem 2 we have $\mathcal{K} \models D'(C_i) \sqsubseteq D'(C_j)$ implies $\mathcal{K} \models \sigma Y.D'(C_i) \sqsubseteq \sigma Y.D'(C_j)$, thus by induction hypothesis we are done. □

Going back to our example, we can, in fact, infer that DAGs of students are also DAGs of persons. Indeed, by applying Theorem 3 and then Theorem 2, we have have that $\mathcal{K} \models student \sqsubseteq person$ implies $\mathcal{K} \models \mu X.emptydag \sqcup (student \sqcap \exists arc.\top \sqcap \forall arc.X) \sqsubseteq \mu X.emptydag \sqcup (person \sqcap \exists arc.\top \sqcap \forall arc.X)$.

## 4.2 Internalizing Assertions

We now show that logical implication in $\mu\mathcal{ALCQ}$ TBoxes (thus also satisfiability of $\mu\mathcal{ALCQ}$ TBoxes) is reducible to unsatisfiability of *a single $\mu\mathcal{ALCQ}$ concept*. To prove this result, we introduce the notions of generated sub-interpretation and sub-valuation[9].

Let $\mathcal{I} = (\Delta^\mathcal{I}, \cdot^\mathcal{I})$ be an interpretation, $\rho$ a valuation on $\mathcal{I}$, and $s \in \Delta^\mathcal{I}$ an individual. We define the interpretation $\mathcal{I}^s = (\Delta^{\mathcal{I}^s}, \cdot^{\mathcal{I}^s})$, and the valuation $\rho^s$ on $\mathcal{I}^s$, as follows:

- $\Delta^{\mathcal{I}^s} = \{s' \in \Delta^\mathcal{I} \mid (s, s') \in (R_1^\mathcal{I} \cup \ldots \cup R_m^\mathcal{I})^*\}$.

- For each atomic role $R_i$, we have $R_i^{\mathcal{I}^s} = R_i^\mathcal{I} \cap (\Delta^{\mathcal{I}^s} \times \Delta^{\mathcal{I}^s})$.

- For each atomic concept $A$, we have $A^{\mathcal{I}^s} = A^\mathcal{I} \cap \Delta^{\mathcal{I}^s}$.

- For each variable $X$, we have $\rho^s(X) = \rho(X) \cap \Delta^{\mathcal{I}^s}$.

We call $\mathcal{I}^s$ *the sub-interpretation of $\mathcal{I}$ generated by $s$*, and $\rho^s$ *the sub-valuation of $\rho$ generated by $s$*.

For generated sub-interpretations and sub-valuations we can state the following lemma.

**Lemma 4** *Let $C$ be a $\mu\mathcal{ALCQ}$ concept. Then for any interpretation $\mathcal{I}$, any valuation $\rho$ on $\mathcal{I}$, and any individual $s \in \Delta^\mathcal{I}$, we have:*

$$\forall t \in \Delta^{\mathcal{I}^s}.\ t \in C_\rho^\mathcal{I} \quad \textit{iff} \quad t \in C_{\rho^s}^{\mathcal{I}^s}.$$

**Proof** Without loss of generality, we consider concepts formed according to the following simplified abstract syntax: $C ::= A \mid \bot \mid \neg C \mid C_1 \sqcap C_2 \mid \exists R.C \mid (\geq n\, R.C) \mid \mu X.C \mid X$.

We prove the result by induction on the number of nested fixpoint constructs. Base case. If in $C$ there are no fixpoint constructs, the thesis can be proven by induction on the formation of $C$.

Inductive case. We assume that the thesis holds for concepts $C$ with $k$ nested fixpoint constructs, and we prove it for concepts $\mu X.C$ with $k + 1$. We recall that, by the Tarski-Knaster Theorem on fixpoints (Tarski, 1955), $t \in (\mu X.C)_\rho^\mathcal{I}$ iff there exists an ordinal $\alpha$ such that $t \in (\mu_\alpha X.C)_\rho^\mathcal{I}$, where $(\mu_\alpha X.C)_\rho^\mathcal{I}$ is defined by transfinite induction as:

- $(\mu_0 X.C)_\rho^\mathcal{I} = \emptyset$

- $(\mu_{\alpha+1} X.C)_\rho^\mathcal{I} = C_{\rho[X/(\mu_\alpha X.C)_\rho^\mathcal{I}]}^\mathcal{I}$

- $(\mu_\lambda X.C)_\rho^\mathcal{I} = \bigcup_{\alpha < \lambda} (\mu_\alpha X.C)_\rho^\mathcal{I}$, if $\lambda$ is a limit ordinal.

Hence we proceed by transfinite induction on ordinals $\alpha$.

Base case of the transfinite induction. $\mu_0 X.C$ is defined as $\bot$, thus trivially we have $t \in (\mu_0 X.C)_\rho^\mathcal{I}$ iff $t \in (\mu_0 X.C)_{\rho^s}^{\mathcal{I}^s}$.

Successor case of the transfinite induction. We want to show that $t \in (\mu_{\alpha+1} X.C)_\rho^\mathcal{I}$ iff $t \in (\mu_{\alpha+1} X.C)_{\rho^s}^{\mathcal{I}^s}$, which reduces to:

$$t \in C_{\rho[X/(\mu_\alpha X.C)_\rho^\mathcal{I}]}^\mathcal{I} \quad \text{iff} \quad t \in C_{\rho^s[X/(\mu_\alpha X.C)_{\rho^s}^{\mathcal{I}^s}]}^{\mathcal{I}^s}. \tag{6}$$

---

9. Together these notions play the same role as that of generated sub-model in modal logics.

To prove this, we start by showing that:

$$t \in C^{\mathcal{I}^s}_{\rho^s[X/(\mu_\alpha X.C)^{\mathcal{I}^s}_{\rho^s}]} \text{ iff } t \in C^{\mathcal{I}^s}_{(\rho[X/(\mu_\alpha X.C)^{\mathcal{I}}_{\rho}])^s}. \tag{7}$$

Notice that the two valuations above may differ only on the value of $X$. If it holds that:

$$t \in X^{\mathcal{I}^s}_{\rho^s[X/(\mu_\alpha X.C)^{\mathcal{I}^s}_{\rho^s}]} \text{ iff } t \in X^{\mathcal{I}^s}_{(\rho[X/(\mu_\alpha X.C)^{\mathcal{I}}_{\rho}])^s}, \tag{8}$$

then by straightforward induction on the formation of $C$ we have that (7) holds as well. Let us prove (8). We can write it as:

$$t \in \rho^s[X/(\mu_\alpha X.C)^{\mathcal{I}^s}_{\rho^s}](X) \text{ iff } t \in (\rho[X/(\mu_\alpha X.C)^{\mathcal{I}}_{\rho}])^s(X),$$

and since $t \in \Delta^{\mathcal{I}^s}$, this reduces to

$$t \in (\mu_\alpha X.C)^{\mathcal{I}^s}_{\rho^s} \text{ iff } t \in (\mu_\alpha X.C)^{\mathcal{I}}_{\rho}.$$

which holds by transfinite inductive hypothesis.

Now, since $C$ contains $k$ fixpoint constructs, by inductive hypothesis on $k$, we have:

$$t \in C^{\mathcal{I}}_{\rho[X/(\mu_\alpha X.C)^{\mathcal{I}}_{\rho}]} \text{ iff } t \in C^{\mathcal{I}^s}_{(\rho[X/(\mu_\alpha X.C)^{\mathcal{I}}_{\rho}])^s}.$$

Hence, considering (6) and (7), it follows that indeed $t \in (\mu_{\alpha+1} X.C)^{\mathcal{I}}_{\rho}$ iff $t \in (\mu_{\alpha+1} X.C)^{\mathcal{I}^s}_{\rho^s}$.

Limit case of the transfinite induction. Let $\lambda$ be a limit ordinal, then $t \in (\mu_\lambda X.C)^{\mathcal{I}}_{\rho}$ iff there exists an ordinal $\alpha < \lambda$ such that $t \in (\mu_\alpha X.C)^{\mathcal{I}}_{\rho}$. By transfinite induction hypothesis, it holds that: $t \in (\mu_\alpha X.C)^{\mathcal{I}}_{\rho}$ iff $t \in (\mu_\alpha X.C)^{\mathcal{I}^s}_{\rho^s}$, and thus:

$$t \in (\mu_\lambda X.C)^{\mathcal{I}}_{\rho} \text{ iff } t \in (\mu_\lambda X.C)^{\mathcal{I}^s}_{\rho^s}.$$

This completes the transfinite induction. So for all ordinals $\alpha$ it holds that:

$$t \in (\mu_\alpha X.C)^{\mathcal{I}}_{\rho} \text{ iff } t \in (\mu_\alpha X.C)^{\mathcal{I}^s}_{\rho^s}.$$

The induction on the nesting of fixpoint constructs is completed as well, hence we have proven the lemma. □

Now we are ready to state the result mentioned above.

**Theorem 5** *Let $\mathcal{K} = \{C_1 \sqsubseteq D_1, \ldots, C_q \sqsubseteq D_q\}$ be a $\mu\mathcal{ALCQ}$ TBox, and $C$ and $D$ two $\mu\mathcal{ALCQ}$ concepts. Then $\mathcal{K} \models C \sqsubseteq D$ if and only if the $\mu\mathcal{ALCQ}$ concept:*

$$\nu X.(\forall R_1.X \sqcap \ldots \sqcap \forall R_m.X \sqcap C_{\mathcal{K}}) \sqcap C \sqcap \neg D \tag{9}$$

*is unsatisfiable, where $R_1, \ldots, R_m$ are all the atomic roles appearing in $\mathcal{K}$, and $C_{\mathcal{K}} = (\neg C_1 \sqcup D_1) \sqcap \ldots \sqcap (\neg C_q \sqcup D_q)$.*

**Proof** If part. By contradiction. Assume that (9) is not satisfiable, and suppose that $\mathcal{K} \not\models C \sqsubseteq D$, i.e., there exists an interpretation $\mathcal{I}$, and a valuation $\rho$ on $\mathcal{I}$, such that $\mathcal{I}$ is a model of $\mathcal{K}$ and $C_\rho^\mathcal{I} \not\subseteq D_\rho^\mathcal{I}$. It follows that, there exists an individual $s \in \Delta^\mathcal{I}$ such that $s \in C_\rho^\mathcal{I}$ and $s \in (\neg D)_\rho^\mathcal{I}$. On the other hand, the fact that $\mathcal{I}$ is a model of $\mathcal{K}$ implies that $(C_\mathcal{K})_\rho^\mathcal{I} = \Delta^\mathcal{I}$, and thus that $(\nu X.(\forall R_1.X \sqcap \ldots \sqcap \forall R_m.X \sqcap C_\mathcal{K}))_\rho^\mathcal{I} = \Delta^\mathcal{I}$. So we have $s \in (\nu X.(\forall R_1.X \sqcap \ldots \sqcap \forall R_m.X \sqcap C_\mathcal{K}) \sqcap C \sqcap \neg D)_\rho^\mathcal{I}$, i.e., (9) is satisfiable, contradicting the hypotheses.

Only If part. Again we proceed by contradiction. Assume $\mathcal{K} \models C \sqsubseteq D$. And suppose that (9) is satisfiable, i.e., there exists an interpretation $\mathcal{I}$, a valuation $\rho$ on $\mathcal{I}$, and an individual $s \in \Delta^\mathcal{I}$, such that $s \in (\nu X.(\forall R_1.X \sqcap \ldots \sqcap \forall R_m.X \sqcap C_\mathcal{K}) \sqcap C \sqcap \neg D)_\rho^\mathcal{I}$.

Now consider the sub-interpretation $\mathcal{I}^s = (\Delta^{\mathcal{I}^s}, \cdot_{\rho^s}^{\mathcal{I}^s})$ and the sub-valuation $\rho^s$ on $\mathcal{I}^s$ generated by $s$. On the one hand, we clearly have that $(C_\mathcal{K})_{\rho^s}^{\mathcal{I}^s} = \Delta^{\mathcal{I}^s}$, hence $\mathcal{I}^s$ is a model of $\mathcal{K}$. On the other hand by Lemma 4 $s \in (\nu X.(\forall R_1.X \sqcap \ldots \sqcap \forall R_m.X \sqcap C_\mathcal{K}) \sqcap C \sqcap \neg D)_{\rho^s}^{\mathcal{I}^s}$, so it follows that $\mathcal{I}^s$ and $\rho^s$ do not satisfy the subsumption $C \sqsubseteq D$, contradicting the hypotheses. $\square$

This result states that satisfiability of $\mu\mathcal{ALCQ}$ concepts and logical implication in $\mu\mathcal{ALCQ}$ TBoxes (and thus of satisfiability of $\mu\mathcal{ALCQ}$ TBoxes) are *not* distinct reasoning tasks. Hence in the following we will limit our attention to concept satisfiability without loss of generality.

# 5. Reasoning with Fixpoints

In this section we concentrate on developing reasoning methods to check for satisfiability concepts involving fixpoints. In particular, we exhibit a correspondence between $\mu\mathcal{ALCQ}$ and a well-known logic of programs, called *modal mu-calculus* (Kozen, 1983; Kozen & Parikh, 1983; Streett & Emerson, 1984, 1989), that has been recently investigated for expressing temporal properties of reactive and parallel processes (Stirling, 1992; Larsen, 1990; Cleaveland, 1990; Winsket, 1989; Dam, 1992).

To get a better insight on the correspondence between the two logics, we first study the sublanguage $\mu\mathcal{ALC}$ obtained from $\mu\mathcal{ALCQ}$ leaving out qualified number restrictions[10]. Then, we study the full logic $\mu\mathcal{ALCQ}$.

## 5.1 Reasoning in $\mu\mathcal{ALC}$

Let us introduce modal mu-calculus formally. Formulae $\Phi, \Psi, \ldots$ of modal mu-calculus are formed inductively from atomic formulae $A, \ldots$ and variables $X, \ldots$ according to the following abstract syntax:

$$\Phi, \Psi ::= A \mid \top \mid \bot \mid \neg\Phi \mid \Phi \wedge \Psi \mid \Phi \vee \Psi \mid \langle a \rangle \Phi \mid [a]\Phi \mid \mu X.\Phi \mid \nu X.\Phi \mid X$$

where $a$ is the generic element of a set of labels $\mathcal{L}$, and every bounded occurrence of every variable $X$ must be in the scope of an even number of negation signs.

---

10. Observe that, in Theorem 5 qualified number restrictions play no role. Hence exactly the same reduction from logical implication to unsatisfiability holds for $\mu\mathcal{ALC}$ as well. This allows us to restrict our attention to satisfiability only.

The semantics of modal mu-calculus is based on the notions of (Kripke) structure and valuation. A *Kripke structure* $\mathcal{M}$ is a triple $(\mathcal{S}, \{\mathcal{R}_i \mid i \in \mathcal{L}\}, \mathcal{V})$, where $\mathcal{S}$ is a set of states, each $\mathcal{R}_i$ is a binary relation on $\mathcal{S}$, and $\mathcal{V}$ is a mapping from atomic formulae to subsets of $\mathcal{S}$. A *valuation* $\rho$ on $\mathcal{M}$ is a mapping from variables to subsets of $\mathcal{S}$. To a Kripke structure $\mathcal{M}$ and a valuation $\rho$ on $\mathcal{M}$, it is associated an extension function $\cdot_\rho^{\mathcal{M}}$ defined inductively as follows:

$$
\begin{aligned}
X_\rho^{\mathcal{M}} &= \rho(X) \subseteq \mathcal{S} \\
A_\rho^{\mathcal{M}} &= \mathcal{V}(A) \subseteq \mathcal{S} \\
\top_\rho^{\mathcal{M}} &= \mathcal{S} \\
\bot_\rho^{\mathcal{M}} &= \emptyset \\
(\neg\Phi)_\rho^{\mathcal{M}} &= \mathcal{S} - \Phi_\rho^{\mathcal{M}} \\
(\Phi \wedge \Psi)_\rho^{\mathcal{M}} &= \Phi_\rho^{\mathcal{M}} \cap \Psi_\rho^{\mathcal{M}} \\
(\Phi \vee \Psi)_\rho^{\mathcal{M}} &= \Phi_\rho^{\mathcal{M}} \cup \Psi_\rho^{\mathcal{M}} \\
(\langle a \rangle \Phi)_\rho^{\mathcal{M}} &= \{s \in \mathcal{S} \mid \exists s'.\ (s, s') \in \mathcal{R}_a \text{ and } s' \in \Phi_\rho^{\mathcal{M}}\} \\
([a]\Phi)_\rho^{\mathcal{M}} &= \{s \in \mathcal{S} \mid \forall s'.\ (s, s') \in \mathcal{R}_a \text{ implies } s' \in \Phi_\rho^{\mathcal{M}}\} \\
(\mu X.\Phi)_\rho^{\mathcal{M}} &= \bigcap\{\mathcal{E} \subseteq \mathcal{S} \mid \quad \Phi_{\rho[X/\mathcal{E}]}^{\mathcal{M}} \subseteq \mathcal{E}\ \} \\
(\nu X.\Phi)_\rho^{\mathcal{M}} &= \bigcup\{\mathcal{E} \subseteq \mathcal{S} \mid \quad \mathcal{E} \subseteq \Phi_{\rho[X/\mathcal{E}]}^{\mathcal{M}}\ \}
\end{aligned}
$$

A formula $\Phi$ is *satisfiable* if there exists a Kripke structure $\mathcal{M}$ and a valuation $\rho$ on $\mathcal{M}$ such that $\Phi_\rho^{\mathcal{M}} \neq \emptyset$.

The following theorem is the basis for the correspondence between $\mu\mathcal{ALC}$ and the modal mu-calculus.

**Theorem 6** *There exists a one-to-one linear-time function $q$ mapping concepts of $\mu\mathcal{ALC}$ to formulae of modal mu-calculus such that: for any $\mu\mathcal{ALC}$ concept $C$, $C$ is satisfiable if and only if $q(C)$ is satisfiable.*

**Proof** We can define $q$ in the following way: $q(A) = A$ (atomic concepts are mapped to atomic formulae), $q(X) = X$, $q(\top) = \top$, $q(\bot) = \bot$, $q(\neg C) = \neg q(C)$, $q(\exists R.C) = \langle R \rangle q(C)$ (atomic roles are mapped to labels), $q(\forall R.C) = [R]q(C)$, $q(\mu X.C) = \mu X.q(C)$, and $q(\nu X.C) = \nu X.q(C)$.

An interpretation $\mathcal{I} = (\Delta^{\mathcal{I}}, \cdot^{\mathcal{I}})$ is equivalent to a Kripke structure $\mathcal{M} = (\mathcal{S}, \{\mathcal{R}_i \mid i \in \mathcal{L}\}, \mathcal{V})$ such that: $\mathcal{S} = \Delta^{\mathcal{I}}$; $\mathcal{L}$ is equal to the set of names of the atomic roles interpreted in $\mathcal{I}$; $\mathcal{R}_R = R^{\mathcal{I}}$ for each atomic role $R$; and $\mathcal{V}(A) = A^{\mathcal{I}}$ for each atomic concept $A$. In addition, a valuation $\rho$ on $\mathcal{I}$ is equivalent to a valuation $\rho'$ on $\mathcal{M}$. Now both the extension function associated with $\mathcal{I}$ and $\rho$ and the extension function associated with $\mathcal{M}$ and $\rho'$ map, respectively, any concept $C$ and the corresponding formula $q(C)$ to the same subset of $\Delta^{\mathcal{I}} = \mathcal{S}$. Hence the thesis follows. $\square$

It follows that we may transfer both decidability and complexity results for the modal mu-calculus (Kozen & Parikh, 1983; Emerson & Jutla, 1988; Safra, 1988) to $\mu\mathcal{ALC}$. Thus, we can immediately state what is the complexity of reasoning with $\mu\mathcal{ALC}$ concepts and $\mu\mathcal{ALC}$ TBoxes.

**Theorem 7** *Satisfiability of $\mu\mathcal{ALC}$ concepts, satisfiability of $\mu\mathcal{ALC}$ TBoxes, and logical implication in $\mu\mathcal{ALC}$ TBoxes are EXPTIME-complete problems.*

**Proof** The satisfiability problem for modal mu-calculus is EXPTIME-complete (Emerson & Jutla, 1988), hence by Theorem 6 and by Theorem 5 the thesis follows. □

## 5.2 Reasoning in $\mu\mathcal{ALCQ}$

Next we exhibit a mapping from $\mu\mathcal{ALCQ}$ concepts to formulae of variant of modal mu-calculus, called *deterministic modal mu-calculus*, which has the same syntax as the modal mu-calculus, but is interpreted over *deterministic Kripke structures*, that is Kripke structures $\mathcal{M} = (\mathcal{S}, \{\mathcal{R}_i \mid i \in \mathcal{L}\}, \mathcal{V})$ in which the relations $\mathcal{R}_i$ are partial functions (Streett & Emerson, 1984).

Let us ignore for a moment the qualified number restriction constructs. Formulae of $\mu\mathcal{ALCQ}$ without qualified number restrictions are, in fact, formulae of the modal mu-calculus, as shown in the previous section. By using a well-known technique developed for propositional dynamic logic (Parikh, 1981), (nondeterministic) modal mu-calculus formulae can be reduced to deterministic modal mu-calculus formulae (Streett & Emerson, 1984), as shown below.

We use the following notations for usual operations on binary relations: $\cdot \circ \cdot$ for *chaining*, $\cdot^*$ for *reflexive transitive closure*, $\cdot^+$ for *transitive closure*, and $\cdot^-$ for *converse*. We also use the following abbreviations:

$$
\begin{array}{lll}
[R^*]\phi & \text{for} & \nu X.(\phi \wedge [R]X) \\
[R^+]\phi & \text{for} & [R][R^*]\phi \\
\langle R^*\rangle\phi & \text{for} & \mu X.(\phi \vee \langle R\rangle X) \\
\langle R^+\rangle\phi & \text{for} & \langle R\rangle\langle R^*\rangle\phi.
\end{array}
$$

The reduction is as follows: in a formula $\Phi$, we recursively replace each subformulae of the form $[R]\phi$ by $[R][(R_{new})^*]\phi$ and each subformulae of the form $\langle R\rangle\phi$ by $\langle R\rangle\langle(R_{new})^*\rangle\phi$, where $R_{new}$ is a new symbol and both $R$ and $R_{new}$ in the resulting formula are interpreted as partial functions. Let us call the resulting formula $\Phi'$, we have that $\Phi$ is satisfiable if and only if $\Phi'$ is satisfiable.

We briefly sketch the reasoning behind the proof of this statement. The if direction is easy: it suffices to observe that if $M^D = (\mathcal{S}^D, \{\mathcal{R}_i^D \mid i \in \mathcal{L}^D\}, \mathcal{V}^D)$ is a model of $\Phi'$, then can transform it into a model $\mathcal{M} = (\mathcal{S}, \{\mathcal{R}_i \mid i \in \mathcal{L}\}, \mathcal{V})$ of $\Phi$ by defining $\mathcal{S} = \mathcal{S}^D$, $\mathcal{L} = \mathcal{L}^D - new$, $\mathcal{R}_R = \mathcal{R}_R^D \circ (\mathcal{R}_{new}^D)^*$, and $\mathcal{V} = \mathcal{V}^D$. The only if direction is as follows. We recall that both nondeterministic and deterministic modal mu-calculus have the tree model property (Streett & Emerson, 1989, 1984): if a formula has a model it has a tree model, i.e., a model having the form of a *tree*[11]. So without loss of generality we can restrict our attention to tree models only. Now there is a one-to-one transformation from tree models $M^T = (\mathcal{S}^T, \{\mathcal{R}_i^T \mid i \in \mathcal{L}^T\}, \mathcal{V}^T)$ of $\Phi$ to (tree) models $M^B = (\mathcal{S}^B, \{\mathcal{R}_i^B \mid i \in \mathcal{L}^B\}, \mathcal{V}^B)$ of $\Phi'$. Indeed, we put $\mathcal{S}^B = \mathcal{S}^T$, $\mathcal{V}^B = \mathcal{V}^T$, $\mathcal{L}^B = \mathcal{L}^T$, and given a state $x \in \mathcal{S}^T$ having as $\mathcal{R}_R^T$-successors $z_1, \ldots, z_l$,[12] we put $(x, z_1) \in \mathcal{R}_R^B$, and $(z_i, z_{i+1}) \in \mathcal{R}_{R_{new}}^B$, for $i = 1, \ldots, l-1$. In this way we have $(x, z_i) \in \mathcal{R}_R^T$ if and only if $(x, z_i) \in \mathcal{R}_R^B \circ (\mathcal{R}_{R_{new}}^B)^*$.[13]

---

11. Given a model of $\Phi$ we get a tree model simply by "unfolding" the original one.

We remark that $M^T$ is required to be a tree because once we get $M^B$ we need to recover the "original" $\mathcal{R}_R^T$-predecessor $x$ of a state $z_i$, namely we need $(\mathcal{R}_R^B \circ (\mathcal{R}_{R_{new}}^B)^*)^-$ to be *a partial function*, otherwise, given a state $z_i$, we would not know which of the various $(\mathcal{R}_R^B \circ (\mathcal{R}_{R_{new}}^B)^*)^-$-successors is its original $\mathcal{R}_R^T$-predecessor $x$, and therefore we would not be able to reconstruct $M^T$ from $M^B$.

By interpreting $R$ and $R_{new}$ as partial functions, it easy to express qualified number restrictions as constraints on the chain of $(R \circ R_{new}^*)$-successors of a state. For example: $(\leq 3 \, R.\phi)$ can be expressed by

$$[R][(R_{new})^*](\neg\phi \vee [(R_{new})^+](\neg\phi \vee [(R_{new})^+](\neg\phi \vee [(R_{new})^+]\neg\phi)))$$

and can be read as "everywhere along the chain $R \circ (R_{new})^*$ there are at most three states where $\phi$ holds", which corresponds exactly to the intended meaning. Similarly $(\geq 3 \, R.\phi)$ can be expressed by

$$\langle R \rangle \langle (R_{new})^* \rangle (\phi \wedge \langle (R_{new})^+ \rangle (\phi \wedge \langle (R_{new})^+ \rangle \phi))$$

and can be read as "somewhere along the chain $R \circ (R_{new})^*$ there are at least three states where $\phi$ holds", which again corresponds exactly to the intended meaning.

The above discussion allows us to state the following result.

**Theorem 8** *There exists a polynomial function $t$ mapping concepts of $\mu\mathcal{ALCQ}$ to formulae of deterministic modal mu-calculus such that: for any $\mu\mathcal{ALCQ}$ concept $C$, $C$ is satisfiable if and only if $u(C)$ is satisfiable.*

**Proof** The function $t$ is defined inductively as follows:

$$
\begin{aligned}
u(A) &= A \\
u(X) &= X \\
u(C_1 \sqcap C_1) &= u(C_1) \wedge u(C_2) \\
u(C_1 \sqcup C_2) &= u(C_1) \vee u(C_2) \\
u(\neg C) &= \neg u(C) \\
u(\mu X.C) &= \mu X.u(C) \\
u(\nu X.C) &= \nu X.u(C) \\
u(\exists R.C) &= \langle R \rangle \langle (R_{new})^* \rangle u(C) \\
u(\forall R.C) &= [R][(R_{new})^*]u(C)
\end{aligned}
$$

where $R_{new}$ is a new role. Finally, $(\leq n \, R.C)$ and $(\geq n \, R.C)$ are mapped to the following formulae:

---

12. We implicitly assume that $M^T$ is a finite branching tree model. This can be done without loss of generality since modal mu-calculus has the finite model property, and hence unfolding a finite model we get a finite branching tree model. Note however that it would suffice to assume $M^T$ to be a countable branching tree model.

13. Note that this construction is similar to the one often used in programming to reduce n-ary trees to binary trees by coding children of a node as the combination of one child and its siblings.

$$u((\leq n\,R.C)) = [R][(R_{new})^*](\neg u(C) \vee [(R_{new})^+](\neg u(C) \vee$$
$$[(R_{new})^+](\ldots(\neg u(C) \vee [(R_{new})^+]\neg u(C))\ldots)))$$

where the number of nested formulae of the form $\neg u(C) \vee [(R_{new})^+]\Phi$ is $n$, and

$$u((\geq n\,R.C)) = \langle R\rangle\langle(R_{new})^*\rangle(u(C) \wedge \langle(R_{new})^+\rangle(u(C) \wedge$$
$$\langle(R_{new})^+\rangle(\ldots(u(C) \wedge \langle(R_{new})^+\rangle u(C))\ldots)))$$

where the number of nested formulae of the form $u(C) \wedge \langle(R_{new})^+\rangle\Phi$ is $n-1$.

$u(C)$ is clearly polynomial in the size of $C$ (under the usual assumption that numbers in $C$ coded in unary). Moreover, following the technique in (Parikh, 1981; Streett & Emerson, 1984) that as been exposed above, it is easy to verify, by induction on the formation of the concept $C$, that the mapping $t$ preserves satisfiability. □

It follows that we may transfer both decidability and complexity results for the deterministic modal mu-calculus (Streett & Emerson, 1984; Emerson & Jutla, 1988; Safra, 1988) to $\mu\mathcal{ALCQ}$. Thus, we can immediately state what is the complexity of reasoning with $\mu\mathcal{ALCQ}$ concepts and $\mu\mathcal{ALCQ}$ TBoxes.

**Theorem 9** *Satisfiability of $\mu\mathcal{ALCQ}$ concepts, satisfiability of $\mu\mathcal{ALCQ}$ TBoxes, and logical implication in $\mu\mathcal{ALCQ}$ TBoxes are EXPTIME-complete problems.*

**Proof** Satisfiability in deterministic modal mu-calculus is an EXPTIME-complete problem (Streett & Emerson, 1984; Emerson & Jutla, 1988; Safra, 1988). Hence by Theorem 8 and Theorem 5 the thesis follows. □

## 6. Discussion and Conclusion

The work presented in this paper stems out from (De Giacomo, 1993), where the basic ideas of introducing explicit fixpoint was first presented, and (De Giacomo & Lenzerini, 1994b), where such idea was further elaborated and $\mu\mathcal{ALCQ}$ was first introduced.

One of the main contributions of this work has been to devise a tight correspondence between description logics with fixpoints and modal mu-calculus. In this respect we remark that, while $\mu\mathcal{ALC}$ corresponds directly to modal mu-calculus, the full $\mu\mathcal{ALCQ}$ corresponds to a variant of modal mu-calculus whose decidability and complexity had not been studied. More precisely, a notion essentially equivalent to that of qualified number restrictions has independently emerged in modal logics, namely that of *graded modalities* (Van der Hoek & de Rijke, 1995; Van der Hoek, 1992; Fattorosi-Barnaba & De Caro, 1985; Fine, 1972). However the combination of fixpoints and graded modalities had not been investigated before in the setting of modal logics. Given the tight correspondence between $\mu\mathcal{ALC}$ and modal mu-calculus, $\mu\mathcal{ALCQ}$ can be considered as *modal mu-calculus augmented with graded modalities.* Hence the results in this paper apply to such a logic as well.

The research reported in this paper bears several similarities with the one on the correspondence between description logics and propositional dynamic logics (Baader, 1991;

Schild, 1991; De Giacomo & Lenzerini, 1994a, 1995; De Giacomo, 1995). In fact what characterize description logics based on propositional dynamic logics are the role constructs for chaining, choice, test, and above all reflexive transitive closure of roles, which is a limited form of fixpoint. Such role constructs can be easily expressed by using the explicit fixpoints introduced here. It suffice to resort to the following equivalences:

$$\exists R_1 \circ R_2.C = \exists R_1.\exists R_2.C$$
$$\exists R_1 \sqcup R_2.C = \exists R_1.C \sqcup \exists R_2.C$$
$$\exists R^*.C = \mu X.(C \sqcup \exists R.X)$$
$$\exists id(D).C = C \sqcap D.$$

Note that $\forall R^*.C = \nu X.(C \sqcap \forall R.X)$. In (Calvanese, De Giacomo, & Lenzerini, 1995) a further implicit form of fixpoint has been advocated, the so called well-founded role construct $wf(R)$. By explicit fixpoints, $wf(R)$ is expressed simply as $\mu X.(\forall R.X)$.

Our proposal of allowing for fixpoint constructs explicitly in the formalism is shared by the study independently carried out by Schild in (Schild, 1994)[14]. The main goal of that work is to study both the expressive power and the computational complexity of subsumption and satisfiability for TBoxes expressed in $\mathcal{ALC}$ (no fixpoint constructs), that allow for *mutually* recursive definitions. To this end, a description logic is defined that corresponds to a variant of the modal mu-calculus in which *mutual fixpoints* are allowed but some restrictions on nested fixpoints are enforced (Vardi & Wolper, 1984). It is well known that mutual fixpoints can be re-expressed by means of nested ones (see, for example, Park, 1976; de Bakker, 1980). As a consequence of this observation it follows that the logic introduced in this paper, is more expressive than the one analyzed in (Schild, 1994) since, on the one hand, it allows nesting of fixpoints without any restriction, on the other hand it makes it possible to state sophisticated forms of cardinality constraints on role fillers by means of qualified number restrictions.

The present work can be extended along several directions. We conclude by outlining two of them.

We already noticed that fixpoint constructs allow for representing not only abstract classes, but also several data structures extensively used in software development. We believe that this characteristic is an important step towards a satisfactory integration of description logics with both traditional and declarative programming systems. Indeed the description logic proposed in this paper provides powerful mechanisms for data structure modeling. In particular, the properties stated in Section 4.1 can be the base to formulate a notion of *parametric concept*[15]. For instance, the expression (named $dag\_of[Z]$)

$$\mu X . emptydag \sqcup (Z \sqcap \exists arc.\top \sqcap \forall arc.X)$$

where $Z$ is a formal parameter, denotes the class of DAGs whose nodes are left unspecified. This class can be used in several ways in the TBox. For example, it can be instantiated by binding the formal parameter to actual parameters, thus getting, say, $dag\_of[student]$, $dag\_of[person]$, etc., which are concepts inheriting the properties of $dag\_of[Z]$.

---

14. In (Schild, 1994) number restrictions are not considered.

15. Note that parametric concepts can be introduced also in simpler logics which do not include fixpoint constructs.

Although $\mu\mathcal{ALCQ}$ is a powerful logic, it lacks the construct for *inverse roles* which is needed for example to correctly capture the notions of (finite) TREE, BINARY-TREE, etc. Indeed, to define the concept of TREE (an EMPTY-TREE is a TREE; a NODE that has at most one parent, some children, and all children are TREEs, is a TREE; nothing else is a TREE) we can write $tree \equiv \mu X . empty\_tree \sqcup (node \sqcap (\le 1\, child^- . \top) \sqcap \exists child. \top \sqcap \forall child. X$ where $child^-$ denotes the inverse of $child$. Notice that the introduction of inverse roles does not pose any difficulty from the semantical point of view; however, its impact on the reasoning method needs to be investigated. More generally, a wide variety of concept constructs can be studied in conjunction with fixpoints. The research on description logics related to propositional dynamic logics in (De Giacomo & Lenzerini, 1994a, 1995; Calvanese et al., 1995; De Giacomo, 1995) may give us hints on how to proceed along this direction.